\documentclass[10pt]{article}

\usepackage{stmaryrd}  
\usepackage{mathrsfs}

\usepackage{verbatim}
\usepackage{url}
\usepackage{amsthm}

\usepackage{amsfonts}
\usepackage{amsmath}
\usepackage{amssymb}
\usepackage{latexsym}
\usepackage{color}
\usepackage[OT2,OT1]{fontenc}

\makeatletter

 \theoremstyle{plain}    
 \newtheorem*{thm*}{Theorem}
 \theoremstyle{plain}    
 \newtheorem{thm}{Theorem}[section]

\newcommand{\calF}{\mathcal{F}}

\newcommand{\calR}{\mathcal{R}}

\newcommand{\calX}{\mathcal{X}}

\def\eps{\epsilon}
\def\Sha{{\ifmmode \text{\fontencoding{OT2}\selectfont SH} 
\else \fontencoding{OT2}\selectfont SH \fi}} 
\DeclareMathOperator{\sgn}{sgn}

\newcommand{\set}[1]{\{#1\}}

\newcommand{\paren}[1]{\left( #1 \right)}
\newcommand{\tlprn}[1]{\left\{ #1 \right\}}
\newcommand{\pred}[1]{\left [\![ #1 \right ]\!]}
\newcommand{\iprod}[2]{\left\langle #1 , #2 \right\rangle}

\newcommand{\ignore}[1]{}

\newcommand{\defeq}{=}

\renewcommand{\vec}[1]{\ensuremath{\text{{\bf\textrm{#1}}}}}

\newcommand{\alf}{\Sigma}
\newcommand{\bw}{w}

\newcommand{\RN}{\R^\N}

\newcommand{\size}{\operatorname{size}}
\newcommand{\poly}{\operatorname{poly}}
\newcommand{\dfa}{\operatorname{DFA}}
\newcommand{\nrm}[1]{\left\Vert #1 \right\Vert}
\newcommand{\oo}[1]{\frac{1}{#1}}
\newcommand{\inv}{^{-1}} 
\newcommand{\abs}[1]{\left| #1 \right|}
\newcommand{\znrm}[1]{\nrm{#1}_0}
\newcommand{\X}{\mathcal{X}}
\newcommand{\C}{\mathcal{C}}

\usepackage{bbm}
\newcommand{\chr}{\boldsymbol{\mathbbm{1}}} 
\renewcommand{\pred}[1]{\chr_{\left\{ #1 \right\}}}

\newcommand{\dirsum}{\oplus}
\newcommand{\supr}[1]{^{(#1)}}

\newcommand{\prs}{\vec{P}}
\newcommand{\pr}[1]{\prs\!\tlprn{#1}}

\newcommand{\beq}{\begin{eqnarray*}}
\newcommand{\eeq}{\end{eqnarray*}}
\newcommand{\beqn}{\begin{eqnarray}}
\newcommand{\eeqn}{\end{eqnarray}}
\newcommand{\ben}{\begin{enumerate}}
\newcommand{\een}{\end{enumerate}}
\newcommand{\bit}{\begin{itemize}}
\newcommand{\eit}{\end{itemize}}

\renewcommand{\a}{\alpha}
\newcommand{\F}{\mathcal{F}}
\newcommand{\N}{\mathbb{N}}
\newcommand{\R}{\mathbb{R}}
\newcommand{\wal}{\bw^{\a}}
\newcommand{\wch}{\bw^{\chi}}

\newcommand{\hide}[1]{}

\title{A Universal Kernel for Learning Regular Languages}

\author{
Leonid (Aryeh) Kontorovich\\
Department of Mathematics\\
Weizmann Institute of Science\\
Rehovot, Israel 76100 
}
\date{}

\begin{document}
\bibliographystyle{plain}
\maketitle

\begin{abstract}

We give a universal kernel that renders all the regular languages 
linearly separable. We are not able to compute this kernel efficiently and 
conjecture that it is intractable, but we do have an efficient $\eps$-approximation.

\end{abstract}

\section{Background}
Since the advent of Support Vector Machines (SVMs), kernel methods
have flourished in machine learning theory
\cite{learning-kernels}. 
Formally, a {\em kernel} is a positive definite function from
$\calX\times\calX$ to $\R$, which, via Mercer's theorem, endows an
abstract set with the structure of a Hilbert space.
Kernels provide both computational and theoretical power. The
so-called {\em kernel trick}, when available, allows us to bypass
computing the explicit embedding $\phi:\calX\to\R^\calF$ in feature
space via the identity $K(x,y)=\iprod{\phi(x)}{\phi(y)}$; this can
lead to a considerable gain in efficiency. On a more conceptual level,
imposing an inner product space structure on an abstract set allows us
to harness the theoretical and computational 
utility
of linear algebra
and convex optimization.

A concrete example where kernel methods provide a palpable advantage
over more direct approaches is that of learning finite automata from
labeled strings. Indeed, the most obvious way to infer a DFA from such
a sample is to build the smallest automaton that accepts all the
positive strings and none of the negative ones. 
A straightforward ``Occam's Razor'' argument \cite[Theorem 2.1]{Kearns97}
shows that with this strategy,
a polynomial (in $1/\epsilon,1/\delta$ and target automaton size)
number of samples is sufficient to ensure a generalization error of no
more than $\epsilon$ with confidence at least $1-\delta$.
Of course, there has to be a catch --
finding the smallest automaton consistent with a set of accepted and
rejected strings was shown to be NP-complete by Angluin
\cite{Angluin78} and Gold \cite{Gold78}; this was further strengthened
in the hardness of approximation result of Pitt and Warmuth
\cite{pitt93minimum}.

In \cite{lcm}, Kontorovich, Cortes and Mohri proposed an alternate
framework for learning regular languages. Strings are embedded in a
high-dimensional space and language induction is achieved by
constructing a maximum-margin hyperplane. This hinges on every
language in a family of interest being {\em linearly separable} under
the embedding, and on the efficient computability of the kernel. This
line of research is continued in \cite{clm07}, where linear
separability properties of rational kernels are investigated.

In this paper,
we give a universal kernel that renders all the regular languages 
linearly separable. 
Any linearly separable language necessarily has a positive margin, and
standard generalization guarantees apply; see \cite{lcm} for details.
We are not able to compute this kernel efficiently and 
conjecture that it is intractable, but we do have an efficient
$\eps$-approximation. Even with these limitations, it appears that the
technique we propose is the first tool to tackle unsupervised learning
of unrestricted regular languages.

\section{Linearly separable concept classes}
\label{sec:linsep-concept}
Let $\C$ be a countable concept class defined over a 
countable
set $\X$. We will say that a
concept $c \in\C$ is {\em finitely linearly separable} if there
exists a mapping $\phi:\X \to \set{0, 1}^\N$ and a weight vector $w
\in \R^\N$,
both with 
\emph{finite support}, i.e.,
$\nrm{w}_0 < \infty$
and
$\nrm{\phi(x)}_0 < \infty$ for all $x\in \X$,
such that
\beq
c = \set{x \in \X: \iprod{w}{\phi(x)} > 0}.
\eeq
The concept class $\C$ is said to be {\em finitely linearly separable}
if all $c \in\C$ are finitely linearly separable under the same mapping
$\phi$.

Note that the condition
$\nrm{\phi(\cdot)}_0 < \infty$ is important; otherwise, we could
define
the {\em embedding by concept}\footnote{
Throughout this paper, we index vectors by integers or members of
other countable sets, as dictated by convenience.} $\phi:\X\to\set{0,1}^\C$
\beq
[\phi(x)]_c &\defeq& \pred{x\in c},
\qquad c\in\C
\eeq
and for any target $\hat c\in\C$,
\beq
w_c &\defeq& \pred{c=\hat c}.
\eeq
This construction trivially ensures that
\beq
\iprod{w}{\phi(x)} &=& \pred{x\in \hat c},
\qquad x\in\X
\eeq
(another reason to require
$\nrm{\phi(\cdot)}_0 < \infty$ is that it automatically makes the kernel
$K(x,y)\defeq\iprod{\phi(x)}{\phi(y)}$ well-defined for all $x,y\in \X$).

Similarly, we disallow 
$\nrm{w}_0 = \infty$ due to the algorithmic impossibility of storing
infinitely many numbers and also because it leads to the trivial
construction, via {\em embedding by instance}:
\beq
[\phi(x)]_u &\defeq& \pred{x=u},
\qquad u\in \X,
\eeq
and for any target $\hat c\in\C$,
\beq
w_u &\defeq& \pred{u\in\hat c}.
\eeq
This again ensures
$\iprod{w}{\phi(x)} = \pred{x\in \hat c}$ without doing anything
interesting or useful.

In light of the examples above, from now on when we speak of linear
separability of a concept class, we shall always assume that $\X$ and
$\C$ are countable and that $w$ and $\phi(\cdot)$ have finite
support. 
An immediate question is whether every 
concept class
is linearly separable in this sense. A positive answer would
require a construction of the requisite $\phi$ given $\X$ and $\C$; a
negative answer would entail an example of $\X$ and $\C$ for which no
such embedding exists.

\section{Every concept class is linearly separable}
In this section we give an affirmative answer to the question raised
in Sec. \ref{sec:linsep-concept}.
\begin{thm}
\label{thm:main}
Every countable 
concept class
$\C$ over a countable 
instance space
$\X$ is linearly
      separable.
\end{thm}
\begin{proof}
Let $\C$ be a countable concept class over the countable instance
space $\X$.
Define two {\em size} functions on $\X$ and $\C$:
$$ \abs{\cdot}:\X\to\N,
\qquad
\nrm{\cdot}:\C\to\N$$
with the property that each has finite level
sets ($\#f\inv(n)<\infty$ for each $n\in\N$); in words, there are at
most finitely many elements of a fixed size. Any countable set has
such a size function.
We will define two auxiliary embeddings, $\chi$ and $\a$, and will
construct the requisite $\phi$ as their direct sum.
For intuition, it is helpful to keep in mind the dual roles of $\X$
and $\C$.
Fix a target $\hat c\in\C$.

Define the {\em embedding by instance}
$\chi:\X\to\set{0,1}^\X$ by
\beq
[\chi(x)]_u &\defeq& \pred{x=u},
\qquad u\in\X;
\eeq
obviously, $\nrm{\chi(x)}_0=1$ for all $x\in\X$.
Define the corresponding hyperplane $\wch\in\R^\X$ by
\beq
[\wch]_u &\defeq& \pred{u\in\hat c}\pred{\abs{u}<\nrm{\hat c}},
\qquad u\in\X;
\eeq
since size functions have finite level sets, we have $\znrm{\wch}<\infty$.
Thus,
\beqn
\nonumber
\iprod{\wch}{\chi(x)} &=&
\sum_{u\in\X} [\wch]_u [\chi(x)]_u \\
\nonumber
&=& \sum_{u\in\X} \pred{u\in\hat c}\pred{\abs{u}<\nrm{\hat c}}\pred{x=u}\\
\label{eq:inst-embed}
&=& \pred{x\in\hat c}\pred{\abs{x}<\nrm{\hat c}}.
\eeqn

Define the {\em embedding by concept} $\a:\X\to\set{0,1}^\C$ by
\beq
[\a(x)]_c &\defeq& \pred{x\in c}\pred{\nrm{c}\leq \abs{x}},
\qquad c\in\C;
\eeq
since size functions have finite level sets, we have
$\nrm{\a(x)}_0<\infty$.
The corresponding hyperplane
$\wal\in\R^\C$ is defined by
\beq
[\wal]_c &\defeq& \pred{c=\hat c},
\qquad c\in\C.
\eeq
Now
\beqn
\nonumber
\iprod{\wal}{\a(x)} &=&
\sum_{c\in\C} [\wal]_c [\a(x)]_c\\
\nonumber
&=& \sum_{c\in\C} 
\pred{c=\hat c} \pred{x\in c}\pred{\nrm{c}\leq \abs{x}} \\
\label{eq:conc-embed}
&=& \pred{x\in\hat c}\pred{\abs{x}\geq\nrm{\hat c}}.
\eeqn

We
define 
the {\em canonical} embedding
$\phi:\X\to\set{0,1}^\N$
as the direct sum of the embeddings by instance and concept:
$$\phi(x)=\chi(x)\dirsum\a(x);$$
note that
$$\znrm{\phi(x)}=\znrm{\chi(x)}+\znrm{\a(x)}<\infty.$$
Similarly, the corresponding hyperplane is the direct sum of the two hyperplanes:
$$\bw = \wch \dirsum \wal;$$
again,
$$\znrm{\bw}=\znrm{\wch}+\znrm{\wal}<\infty.$$
Combining (\ref{eq:inst-embed}) and (\ref{eq:conc-embed}), we get
\beq
\iprod{\bw}{\phi(x)} &=& \iprod{\wch}{\chi(x)} + \iprod{\wal}{\a(x)}\\
&=& 
\pred{x\in\hat c}
\pred{\abs{x}<\nrm{\hat c}}
+
\pred{x\in\hat c}
\pred{\abs{x}\geq\nrm{\hat c}}\\
&=& \pred{x\in \hat c}
\eeq
which shows that $\bw$ is indeed a linear separator (with finite
support) for 
$\hat c$.
\end{proof}

\section{Universal regular kernel}
To apply Theorem \ref{thm:main} to regular languages (over a fixed
alphabet $\Sigma$), we observe that
the DFAs
are a countable concept class 
$\calR=\cup_{n\geq1}\dfa(n)$ over
$\X=\Sigma^*$,
where 
$\dfa(n)$ is the set of all DFAs on $n$ states.
Denoting 
by
$\nrm{A}$ the number of states in $A\in\calR$, we see that
$\nrm{\cdot}$ 
is a valid size function on
$\calR$. A natural size function on $\Sigma^*$ is string length,
denoted by $\abs{\cdot}$. With these two size functions, 
Theorem \ref{thm:main} furnishes an embedding
$\phi:\calR\to\set{0,1}^\N$ that renders all regular languages
linearly separable. To get a better feel for this embedding, let us
compute its associated kernel
\beq
K(x,y) &=& \iprod{\phi(x)}{\phi(y)} \\
&=& \pred{x=y} + \sum_{n=1}^{\min\set{|x|,|y|}} K_n(x,y)
\eeq
where
\beqn
\label{eq:kndef}
K_n(x,y) &=& 
\sum_{A\in\dfa(n)} \pred{x\in L(A)}\pred{y\in L(A)}
.
\eeqn
In other words, $K_n(x,y)$ counts the number of $n$-state DFAs that accept
both $x$ and $y$. 
By \cite[Theorem 6]{lcm},
an immediate consequence of this construction is
that 
every regular language $L$ can be represented by some {\em support
strings}
$\set{s_i\in\alf^*:1\leq i\leq m}$
with weights
$\alpha\in\R^m$:
\beq
L &=& \tlprn{x\in\alf^*: \sum_{i=1}^m \alpha_iK(s_i,x)>0}.
\eeq

\section{Computing $K_n$}
Since the summation in (\ref{eq:kndef}) involves a super-exponential
number of terms, brute-force evaluation is out of the question. Though
we consider the complexity of $K_n$ to be a likely candidate for
$\#$P-complete, we have no proof of this; there is also the hope that
the symmetry in the problem will enable a clever efficient
computation.

In the meantime, we must resort to a Monte Carlo simulation. For
$n>0$ and $x,y\in\Sigma^*$, define $P_n(x,y)$ to be the fraction of
all the DFAs on $n$ states that accept both $x$ and $y$. Thus, $0\leq
P_n(x,y)\leq 1$, and computing this quantity is tantamount to
computing 
$K_n(x,y)
=P_n(x,y)\abs{\dfa(n)}
$. Now it is a simple matter to generate $n$-state
DFAs uniformly
at random. Let
$\set{A_i:1\leq i\leq m}$ be such an independent sample of $m$-state
DFAs, and compute the approximation to $P_n(x,y)$:
\beq
\hat P_n(x,y) &=& \oo m \sum_{i=1}^m \pred{x\in L(A_i)}\pred{y\in L(A_i)}.
\eeq
Then, by Chernoff's bound, we have
\beq
\pr{\abs{\hat P_n(x,y) - P_n(x,y)}>\eps P_n(x,y)  }
&\leq& 2\exp(-\eps^2m P_n(x,y)/3 ),
\eeq
meaning that with probability at least 
$1-2\exp(-2\eps^2m P_n(x,y))$, we have
\beq
(1-\eps) \hat K(x,y) \leq K(x,y) \leq (1+\eps) \hat K(x,y),
\eeq
where 
$\hat K_n(x,y) 
=
\hat 
P_n(x,y)
\abs{\dfa(n)}
$.
Thus, we need
\beq
m &\geq& 3\log(2/\alpha) \over \eps^2 P_n(x,y)
\eeq
sampling steps to have an $\eps$-approximation to $K(x,y)$ with
probability at least $1-\alpha$.

It remains to lower-bound $P_n(x,y)$; if it turns out to be
exponentially small in automaton size $n$, the $\eps$-approximation
will require exponentially many steps. Fortunately, this does not
happen:
\begin{thm}
For all $n\geq1$, for all $x,y\in\alf^*$, we have
\beq
\oo4 \leq P_n(x,y) \leq \oo2.
\eeq
\end{thm}
\begin{proof}
The upper bound is simple -- it follows from the fact that $K_n(x,x) =
\oo2 
\abs{\dfa(n)}
$. Indeed, 
for any $x\in\alf^*$, for every
$A^+\in\dfa(n)$ that accepts $x$ there is exactly one 
$A^-\in\dfa(n)$ that does not (obtained by changing the state in
which $A^+$ ends up after reading $x$ from accepting to
non-accepting). The upper bound follows from the obvious relation
$K_n(x,y)\leq K_n(x,x)$ for all $x,y\in\alf^*$.

To prove the lower bound, take the ``worst'' case where $x,y\in\alf^*$
are such that every $A\in\dfa(n)$ has
$\delta(q_0,x)\neq\delta(q_0,y)$. In other words, no automaton ends up in
the same state after reading $x$ and as it does after reading
$y$.
Since every
state is independently chosen to be accepting or not 
with equal probability, exactly one-fourth of all $A\in\dfa(n)$
will accept both $x$ and $y$. Clearly, this fraction will be higher if
we allow some automata to end up in the same state upon reading $x$
and $y$.
\end{proof}

This means that if we run the (very simple and efficient) simulation
algorithm for $m=12\eps^{-2}\log(2/\alpha)$ steps, we will have an
$\eps$-approximation to $K_n(x,y)$ with probability at least
$1-\alpha$.

\section{Conclusion}
Many fascinating questions arise naturally around the kernel $K_n$
that we defined: Is it (or any other universal regular kernel)
efficiently computable? How can one efficiently recover the automaton
from the hyperplane? Can quantitative margin bounds be obtained
(perhaps in terms of automaton size)? These questions hold potential
for promising future research.

\section*{Acknowledgments}
Various key concepts were crystallized during 
the many valuable discussions
with
Corinna Cortes and Mehryar Mohri.
Thanks also to
Jeremiah Blocki,
Avrim Blum,
Manuel Blum,
Daniel Golovin, 
Nati Linial
and
Noam Zeilberger
for helpful and insightful input.

\bibliography{univker}
\end{document}